\def\year{2022}\relax
\documentclass[letterpaper]{article}
\usepackage{aaai22} 
\usepackage{times} 
\usepackage{helvet} 
\usepackage{courier} 
\usepackage[hyphens]{url} 
\usepackage{graphicx} 
\usepackage{graphicx} 
\usepackage{natbib} 
\usepackage{caption} 
\DeclareCaptionStyle{ruled}%
{labelfont=normalfont,labelsep=colon,strut=off}
\usepackage{algorithm}
\usepackage{algorithmic}
\usepackage{booktabs}
\usepackage{newfloat}
\usepackage{listings}
\usepackage{todonotes}
\floatstyle{ruled}
\newfloat{listing}{tb}{lst}{}
\floatname{listing}{Listing}

\title{Supervising Model Attention with Human Explanations for \\  Robust Natural Language Inference}
\author{
    Joe Stacey\textsuperscript{\rm 1},
    Yonatan Belinkov\textsuperscript{\rm 2},
    Marek Rei\textsuperscript{\rm 1}
}
\affiliations{
    \textsuperscript{\rm 1}Imperial College London\\

    \textsuperscript{\rm 2}Technion -- Israel Institute of Technology

    j.stacey20@imperial.ac.uk, belinkov@technion.ac.il, marek.rei@imperial.ac.uk
}

\usepackage{bibentry}

\begin{document}

\maketitle

\begin{abstract}
Natural Language Inference (NLI) models are known to learn from biases and artefacts within their training data, impacting how well they generalise to other unseen datasets. Existing de-biasing approaches focus on preventing the models from learning these biases, which can result in restrictive models and lower performance. We instead investigate teaching the model how a human would approach the NLI task, in order to learn features that will generalise better to previously unseen examples.
Using natural language explanations, we supervise the model's attention weights to encourage more attention to be paid to the words present in the explanations, significantly improving model performance. Our experiments show that the in-distribution improvements of this method are also accompanied by out-of-distribution improvements, with the supervised models learning from features that generalise better to other NLI datasets. 
Analysis of the model indicates that human explanations encourage increased attention on the important words, with more attention paid to words in the premise and less attention paid to punctuation and stop-words.
\end{abstract}

\section{Introduction}

Natural Language Inference (NLI) models predict the relationship between a premise and hypothesis pair, deciding whether the hypothesis is entailed by the premise, contradicts the premise, or is neutral with respect to the premise. While NLI models achieve impressive in-distribution performance, they are known to learn from dataset-specific artefacts, impacting how well these models generalise on out-of-distribution examples \cite{gururangan-etal-2018-annotation, tsuchiya-2018-performance, poliak-etal-2018-hypothesis}. De-biasing efforts to date have successfully improved out-of-distribution results, but mostly at the expense of in-distribution performance \cite{belinkov-etal-2019-dont,karimi-mahabadi-etal-2020-end,sanh2020learning}.

\begin{figure}[t!]
    \includegraphics[height=95pt]{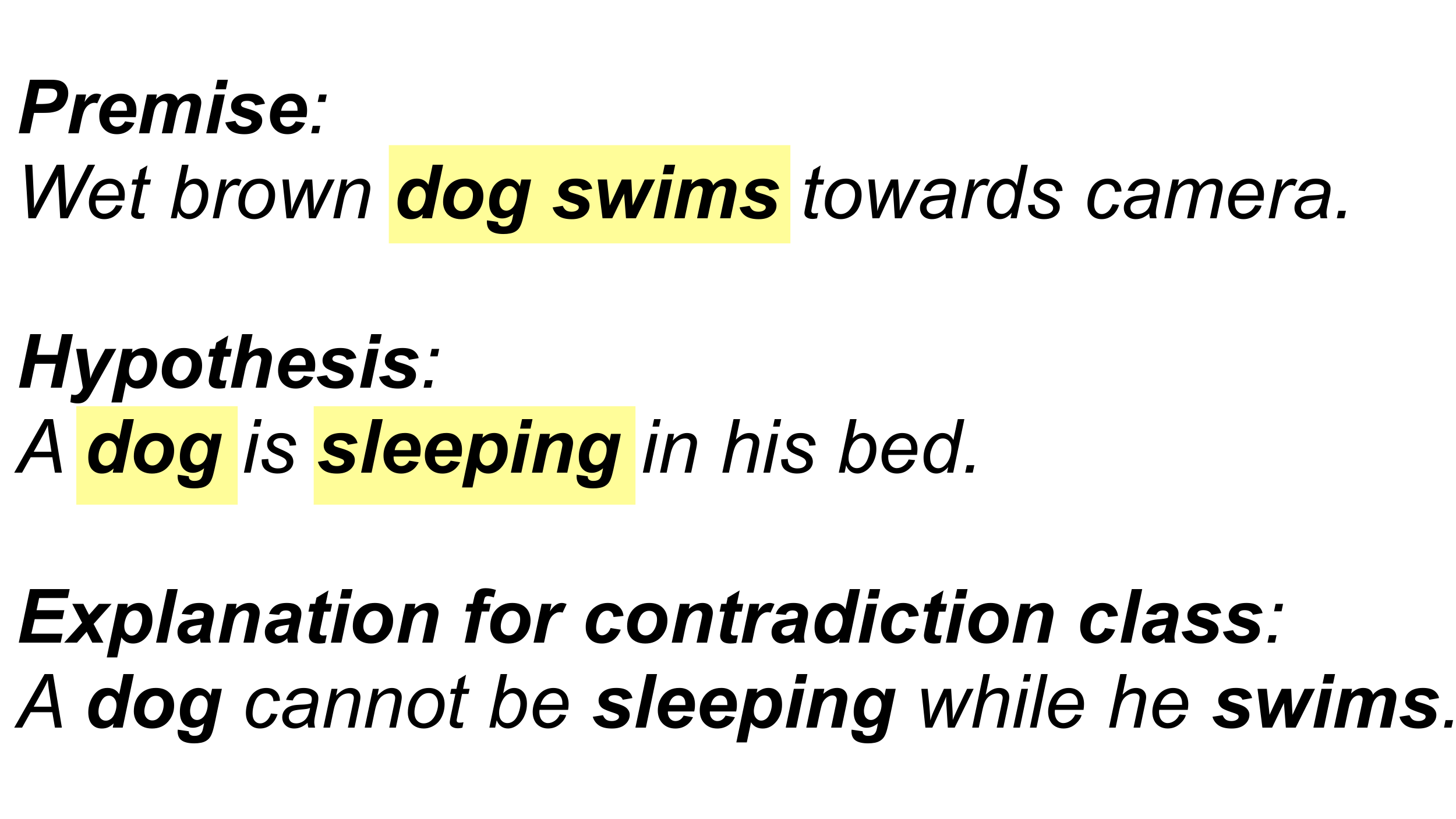} 
    \caption{An example of using a free text explanation to identify important words in the premise and hypothesis. In this case the words \textit{dog}, \textit{sleeping} and \textit{swims} have been identified from the explanation.} \label{example}
\end{figure}

While most previous work creating more robust NLI models has focused on preventing models learning from biases or artefacts in their datasets (more details in the Related Work section), we take a different approach. We aim to use information about how humans approach the task, training with natural language explanations in the e-SNLI dataset \cite{camburu2018esnli} to create more robust models.

Human explanations have been found to improve performance on a range of tasks \cite{rajani-etal-2019-explain, andreas-etal-2018-learning,
mu-etal-2020-shaping, liang-etal-2020-alice}; however, this has largely not been the case in NLI \cite{hase2021models, kumar-talukdar-2020-nile, camburu2018esnli}. Generating human explanations from e-SNLI has been found to improve model performance \citep{zhao2020lirex}, but this process is highly computationally expensive and the in-distribution improvements are accompanied by a reduction in out-of-distribution performance. We aim to address both issues, proposing a simple and efficient method for using explanations to improve model robustness while also improving in-distribution performance.

We investigate multiple approaches to incorporate these human explanations. 
Firstly, we introduce an additional loss term to encourage the model to pay more attention to words in the explanation, supervising the attention from the [CLS] token in the existing model self-attention layers. Additionally, we introduce another attention layer on top of the model and supervise its weights. We also adapt a further attention-based approach for incorporating explanations as proposed by \citet{pruthi2020evaluating}, testing whether this method also improves performance and model robustness for NLI. Each approach considers the most important words in the hypothesis and premise based on the e-SNLI human explanations (see Figure \ref{example}).

To summarise our contributions:
1) We propose a method for supervising with human explanations that provides significant improvements on both in-distribution and out-of-distribution NLI datasets.
2) We show that when combined with DeBERTa \cite{he2021deberta}, this approach achieves a new state-of-the-art result for SNLI \cite{bowman-etal-2015-large}.
3) We show that the model attention weights can effectively predict which words will appear in the explanations, reaching the same performance as prior work that focuses on this task.
4) Finally, we show that training with human explanations encourages the model to pay more attention to important words in the premise and focus less on stop-words in the hypothesis, helping to mitigate the hypothesis-only bias of NLI systems \citep{gururangan-etal-2018-annotation}.\footnote{https://github.com/joestacey/NLI\_with\_a\_human\_touch}

\section{Related Work}
\subsection{Training NLI Models with Explanations}

Most work to date has found that training with NLI explanations does not translate into either in-distribution or out-of-distribution improvements \cite{camburu2018esnli, kumar-talukdar-2020-nile, hase2021models}.
\citet{camburu2018esnli} implement two approaches for incorporating the model explanations: using an \textit{Explain then Predict} approach which generates an explanation and uses it to predict the class, and also predicting both the NLI class and generating the explanation from the same vector of features.
Neither of these approaches significantly improved performance in-distribution or out-of-distribution on the MNLI dataset.

\citet{hase2021models} use a retrieval-based approach for incorporating the e-SNLI explanations, retrieving the top explanations for a hypothesis and premise pair and combining the sentences with the retrieved explanations. They conclude that the \mbox{e-SNLI} dataset does not meet the six preconditions for their retrieval approach to improve performance, with these conditions including how explanations need to be sufficiently relevant across data points.

\citet{kumar-talukdar-2020-nile} generate explanations specific to each class, using these explanations along with the premise and hypothesis to predict the NLI class. This corresponds to a drop in performance both in-distribution and out-of-distribution \cite{kumar-talukdar-2020-nile}.
\citet{zhao2020lirex} also generate explanations for each class, first predicting which of the words in a hypothesis are relevant given the class, training with the highlighted words in e-SNLI. Explanations are then generated based on these annotated hypotheses. While this approach did improve in-distribution performance, out-of-distribution performance did not improve. This process involved training a pipeline of three RoBERTa \citep{liu2019roberta} models and a GPT2 \cite{radford2019language} model, with the performance of this pipeline compared to the performance of a single RoBERTa baseline model.

Unlike the prior work, we aim to show how training with human explanations can improve out-of-distribution performance. We also aim to show that in-distribution improvements are possible
within a single model, without requiring a pipeline of models,
and that these in-distribution and out-of-distribution benefits can be achieved simultaneously.

\subsection{Training with Explanations Beyond NLI}

\citet{pruthi2020evaluating} introduce a teacher-student framework for training with explanations, finding that attention-based approaches are the most effective way to improve performance on sentiment analysis and question answering tasks. For sentiment analysis this involved supervising the attention from the [CLS] token. Attention-based methods to incorporate explanations have also been found to improve performance on hate speech detection \cite{mathew2020hatexplain}. 

Closest to our work, \citet{pruthi2020evaluating} supervise the average attention weights across all of a model's attention heads, whereas we identify which specific heads benefit the most from the supervision and then supervise these heads individually. Their method uses KL-Divergence as an auxiliary loss, while we found mean squared error to perform better when supervising attention. Moreover, \citet{pruthi2020evaluating} do not consider out-of-distribution performance, which is the focus of our work, and do not use free-text explanations, while we incorporate explanations either as free-text explanations or in the form of highlighted words.

\citet{pruthi2020evaluating} train with up to 1,200 and 2,500 examples across two tasks, while we train with a large corpus of 550,152 training observations. As there is more benefit from the explanations when training with fewer examples \citep{pruthi2020evaluating}, it is also not clear whether the improvements will translate to a dataset of this scale. \citet{pruthi2020evaluating} also investigate training with explanations for sentiment analysis and question answering tasks, whereas we train with explanations for NLI, a task where most prior work finds that explanations do not improve performance \cite{hase2021models, kumar-talukdar-2020-nile, camburu2018esnli}. We investigate the performance from adapting the method proposed by \citet{pruthi2020evaluating} to NLI, in addition to comparing this with the improvements from our two proposed approaches.

More widely, explanations have improved performance on a range of domains, including commonsense reasoning \cite{rajani-etal-2019-explain}, relation extraction \cite{murty-etal-2020-expbert} and visual classification tasks \cite{liang-etal-2020-alice, mu-etal-2020-shaping}. Prior work focuses on finding in-distribution improvements rather than considering model robustness, whereas we find that the largest impact from training with model explanations can be the corresponding improvements in model robustness.

\subsection{Creating More Robust NLI Models}

Previous work on creating more robust NLI models has focused on preventing models learning from artefacts (or \textit{biases}) in their training data. 
The most common strategy for mitigating biases within NLI is by creating a weak model to intentionally learn a bias, then encouraging a target model to have low similarity to this weak model \cite{he-etal-2019-unlearn, clark-etal-2019-dont,karimi-mahabadi-etal-2020-end, utama-etal-2020-towards, sanh2020learning, liu-etal-2020-empirical, clark-etal-2020-learning} or to use the weak model to weight training observations \cite{clark-etal-2019-dont, utama-etal-2020-towards, liu-etal-2020-empirical}.
\begin{figure}[ht!]
    \begin{center}
    \includegraphics[width=\columnwidth]{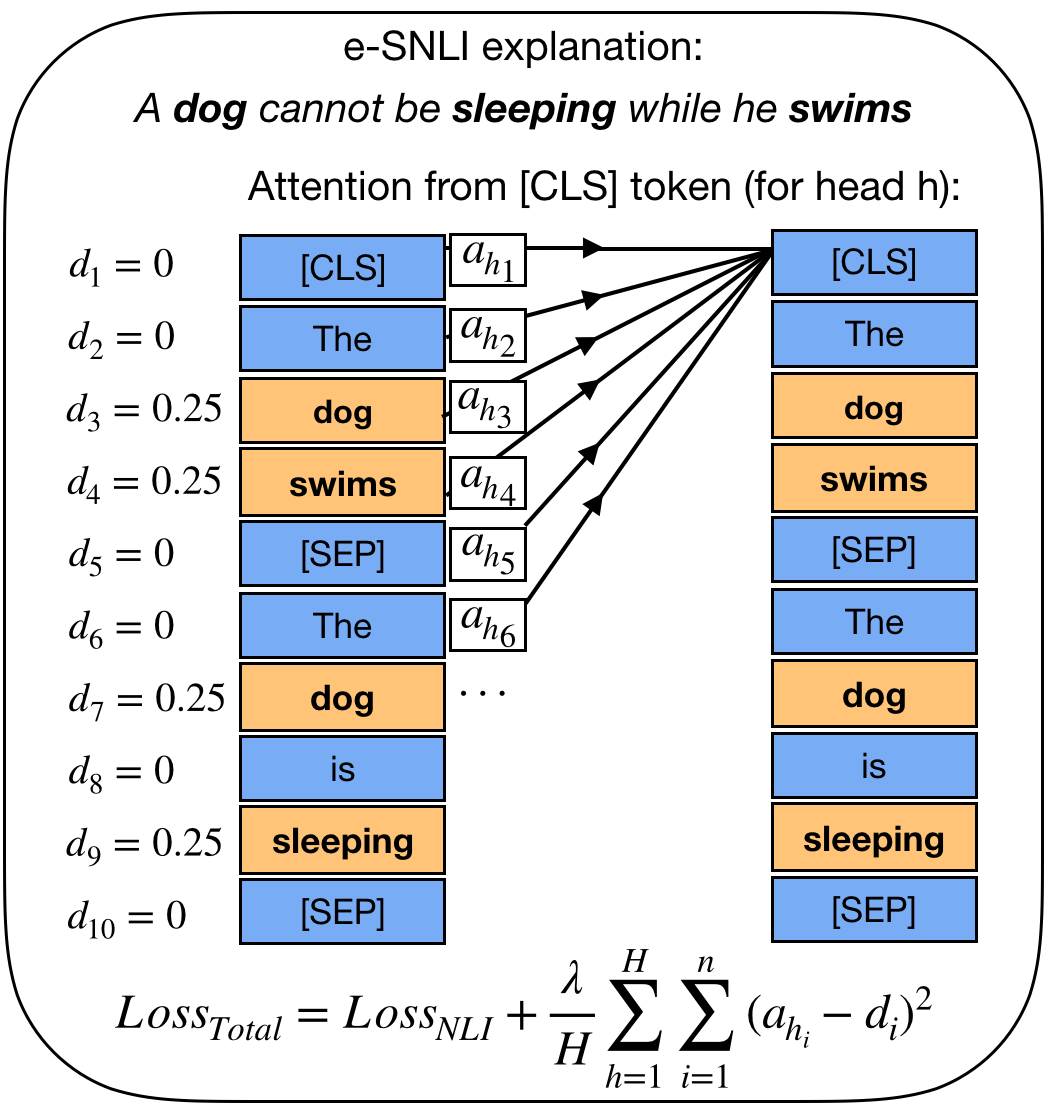}
    \end{center}
    \caption{An example of how the attention loss is calculated when supervising an existing self-attention layer.} \label{model_schematic}
\end{figure}

Other strategies to prevent models learning from artefacts include using adversarial training with gradient reversal to mitigate the hypothesis-only bias \citep{belinkov-etal-2019-dont,belinkov-etal-2019-adversarial, stacey-etal-2020-avoiding}, using data-augmentation \cite{min-etal-2020-syntactic, minervini-riedel-2018-adversarially}, fine-tuning on minority examples \cite{yaghoobzadeh-etal-2021-increasing}, gradient supervision with counterfactual examples \cite{teney2020learning}, multi-task learning \cite{tu-etal-2020-empirical} or creating compressed representations to remove irrelevant information \cite{mahabadi2021variational}.  We take a new and different approach, encouraging models to learn from how humans would approach the task.

\section{Attention Supervision Method}

The e-SNLI explanations \cite{camburu2018esnli} were created by asking Amazon Mechanical Turk annotators why each hypothesis and premise had their given label. The explanations take the form of either free text explanations, or highlighted words in the premise and hypothesis that annotators believe are important. Based on these explanations we create labels $E = \{e_i\}_{i=1}^n$ for each observation, with $e_i$ taking values of either 0 or 1 to indicate whether a token is relevant to a human explanation, and $n$ being the number of tokens in the NLI sentence pair. For free-text explanations, $e_i$ has a value of 1 if its corresponding token is from a word present in the explanation, otherwise the value is 0. For the highlighted words, $e_i$ has a value of 1 if the corresponding word in the premise or hypothesis has been highlighted by the annotator. For the free text explanations we exclude stop-words, whereas highlighted stopwords are selected.\footnote{Performing the matching based on free text would return many incorrect stop-words, whereas using the highlights allows us to focus specifically on the ones that the annotators have selected.}

These explanations are only used during training, whereas during testing the model predicts the NLI class based on the hypothesis and premise alone.

\subsection{Supervising Self-Attention Layers}

To supervise the model's attention weights we create a desired distribution $D = \{d_i\}_{i=1}^n$ of attention values, normalizing the $e_i$ values to sum to 1:
\[d_i = \frac{e_i}{\sum_{k=1}^{n} {e_k}}\]

We supervise the [CLS] attention weights in the final self-attention layer of a transformer model, introducing a second loss term to encourage assigning more attention to words in the human-annotated explanations (see Figure \ref{model_schematic}). We supervise the attention weights in the final self-attention layer as we find this performs better than supervising previous layers. Where $a_{h_i}$ denotes the attention weights for a given attention head, the total loss is defined as:
\[Loss_{Total} =  Loss_{NLI} + \frac{\lambda}{H} \sum_{h=1}^{H}\sum_{i=1}^{n}(a_{h_i} - d_i)^2\]
\noindent where $Loss_{NLI}$ is the main cross-entropy loss for the NLI task, $H$ is the number of heads being supervised and $\lambda$ is a hyper-parameter weighting the attention component of the model loss. The attention values for a given head $a_{h_i}$ are defined as:
\[a_{h_i} =  \frac{\exp{(q^T_{h_{CLS}}}k_{h_i} / \sqrt{d_k})} {\sum_{j=1}^{n}\exp{(q^T_{h_{CLS}}}k_{h_j} / \sqrt{d_k})}\]
Where $q_{h_{CLS}}$ represents the CLS query vector for the head, $k_{h_i}$ are the key vectors for the other tokens in the sentence and $d_k$ is the dimensionality of the key vectors.

\subsection{Selecting Attention Heads for Supervision}
As the attention heads can have different roles \cite{clark-etal-2019-bert,vig-belinkov-2019-analyzing}, when supervising an existing self-attention layer we investigate how many and which heads should be supervised. We supervise each attention head in turn to investigate which heads benefit the most from the supervision. We then choose the top \textit{K} heads for supervision, where \textit{K} is a hyper-parameter tuned across the values $\{1, 3, 6, 9, 12\}$ using 5 random seeds for each condition. This greedy approach does not guarantee finding the optimal subset of heads, but it is more efficient than trying all subsets. By introducing this approach to selectively supervise the attention heads, the model can benefit from the explanation supervision while also allowing for diversity between the roles of the supervised and unsupervised attention heads.
\begin{table*}[!t]
\begin{center}
\begin{tabular}{rcccccccccc}
\toprule
%
%
%
 & {\bf Dev} & {\bf Test} & {\bf Hard} & {\bf MNLI mi} & {\bf MNLI ma} & {\bf ANLI} & {\bf HANS }\\ 

\midrule
BERT baseline & 90.05 & 89.77 & 79.36 & 72.52 & 72.28 & 31.81 & 56.83 \\
\midrule
Ours (extra layer) & 90.40 & 90.09 & 79.96 & 73.03 & 73.10 & 31.47 & 57.85\\
Improvement & \textbf{+0.35}$\dagger \ddagger$  &\textbf{+0.32}$\dagger \ddagger$  & \textbf{+0.60}$\dagger \ddagger$  & \textbf{+0.51}$\dagger$  & \textbf{+0.82}$\dagger \ddagger$  & -0.34 & +1.02\\
\midrule
Ours (existing attention) & 90.45 & 90.17 & 80.15 & 73.36 & 73.19 & 31.41 & 58.42 \\
Improvement & \textbf{+0.40}$\dagger \ddagger$ & \textbf{+0.40}$\dagger \ddagger$ & \textbf{+0.79}$\dagger \ddagger$ & \textbf{+0.84}$\dagger \ddagger$ & \textbf{+0.91}$\dagger \ddagger$ & -0.40 & \textbf{+1.59} $\dagger$ \\

\bottomrule
\end{tabular}

\end{center}
\caption{Average accuracy across 25 random seeds, evaluated on: SNLI-dev, SNLI-test, SNLI-hard, MNLI mismatched (MNLI mi), MNLI matched (MNLI ma), ANLI (R1, R2 and R3) and HANS. Ours (extra layer) involves creating and supervising an additional attention layer on top of the model, while Ours (existing attention) involves supervising 3 heads of an existing self-attention layer. Significant results with P-values less than 0.05 are shown in bold and with a $\dagger$. $\ddagger$ indicates results that are statistically significant after applying a Bonferroni correction factor of 7 for each dataset tested.}
\label{significance_testing}
\end{table*}
\subsection{Supervising an Additional Attention Layer}
Instead of supervising an existing self-attention layer in the model, an additional attention layer can also be created using the sequence representations  $\{h_i\}$ from the transformer model. Using an architecture similar to \citet{DBLP:conf/aaai/ReiS19}, we define our unnormalised attention values $\widetilde{a_i}$ as:
\[\widetilde{a_i} = \sigma{(W_{h2}(\tanh{(W_{h1} h_i + b_{h1})})+b_{h2})} \]
where $W_{h1}$ and $W_{h2}$ are trainable parameters along with their respective bias terms. We supervise the normalized attention weights $a_i$: \[a_i = \frac{\widetilde{a}_i}{\sum_{k=1}^{n}\widetilde{a}_k}\]
These weights are used to create a new representation $c$:

\[c = \sum_{i=1}^{n}a_{i}h_i\]

Finally, a linear classifier and softmax are applied to this representation to predict the class. $Loss_{Total}$ is the same as described previously, using the single attention head.

\subsection{Experimental Setup and Evaluation}
The attention supervision was implemented with BERT \cite{devlin-etal-2019-bert} and DeBERTa \cite{he2021deberta}, the latter using disentangled matrices on content and position vectors to compute the attention weights. We use DeBERTa to assess whether our proposed approach can improve on current state of the art results. $\lambda$ was chosen based on performance on the validation set, trying values in the range $[0.2, 1.8]$ at increments of $0.2$. For our BERT model the best performing $\lambda$ is 1.0, equally weighting the two loss terms, whereas for DeBERTa this value was 0.8.

The robustness of the model is assessed by significance testing on the MultiNLI matched and mismatched validation sets \cite{williams-etal-2018-broad}, and the ANLI \cite{nie-etal-2020-adversarial}, SNLI-hard \cite{gururangan-etal-2018-annotation} and HANS \cite{mccoy-etal-2019-right} challenge sets, using a two-tailed t-test to assess significant improvements from the baseline. HANS contains examples where common syntactic heuristics fail, while SNLI-hard is created from the SNLI test set with examples that a hypothesis-only model has misclassified. ANLI is created using a human-in-the-loop setup to create intentionally challenging examples. The SNLI dev and test set are considered in-distribution, while HANS, ANLI, SNLI-hard and the MNLI mismatched and matched datasets are considered out-of-distribution. 

\section{Experiments}

\subsection{Performance in and out of Distribution} 

The experiments show that supervising the attention patterns of BERT based on human explanations simultaneously improves both in-distribution and out-of-distribution NLI performance (Table \ref{significance_testing}). When supervising an existing self-attention layer, in-distribution accuracy on the SNLI test set improves by $0.4\%$. The hard subset of this set, SNLI-hard, has a larger improvement of 0.79\%, showing that the human explanations provide the most benefit for the hardest SNLI examples. The improvements in SNLI-test and SNLI-hard are significant, with p-values less than $10^{-8}$. Moreover, out-of-distribution performance improves on both of the MNLI validation sets and on HANS, with accuracy improvements of 0.84\%, 0.91\% and 1.59\% respectively (see bottom half of Table \ref{significance_testing}). We do not see improvements on the highly-challenging ANLI dataset, where multiple sentences were used for each premise.

To ensure that these improvements are not simply caused by regularization from supervising the attention weights, we create a randomised baseline by shuffling our desired distribution \textit{D}, doing this separately for the premise and hypothesis. This highlights the effect of the supervision but without the additional information from the explanations. We find that this randomised baseline performs worse than the baseline with no supervision (89.50\% accuracy on SNLI-test), with lower performance also seen on SNLI-hard (78.84\%) and the MNLI datasets (71.5\% and 71.23\%).
\begin{table*}[!t]
\begin{center}
\begin{tabular}{rcccccccccc}
\toprule


 & {\bf SNLI} & {\bf $\Delta$} &
 {\bf MNLI} & {\bf $\Delta$} &
 {\bf SNLI-hard} & {\bf $\Delta$} & \textbf{Params.}\\ 

\midrule
BERT Baseline & 89.77 & & 72.40 & & 79.36 & & 109m\\
\midrule
LIREx-adapted & \textbf{90.79} & \textbf{+1.02}$\dagger$ & 71.55 & -0.85$\dagger$ & 79.39 & +0.03 & 453m\\
Pruthi et al-adapted. & 89.99 & +0.22$\dagger$ & 73.27 & +0.87$\dagger$ & 79.90 & +0.54$\dagger$ & 109m\\
\midrule
Ours (extra layer) & 90.09 & +0.35$\dagger$ & 73.06 & +0.67$\dagger$&79.96& +0.60$\dagger$ & 109m\\

Ours (existing attention) & 90.17 & +0.40$\dagger$ & \textbf{73.28} & \textbf{+0.88}$\dagger$&\textbf{80.15}& \textbf{+0.79}$\dagger$ & 109m\\

\bottomrule
\end{tabular}

\end{center}
\caption{Accuracy improvements compared to previous work, adapting \citet{pruthi2020evaluating} for NLI and adapting LIREx \citep{zhao2020lirex} to use BERT models instead of the three RoBERTa models in its pipeline. $\dagger$ indicates statistically significant results compared to the baseline. Our methods and the \citet{pruthi2020evaluating} method were tested over the same 25 random seeds, while the highly computationally expensive LIREx-adapted approach was evaluated over 5 random seeds. }
\label{benchmarking_results}
\end{table*}

When introducing an additional attention layer, the model with this extra layer does not outperform the baseline if the additional layer is not supervised. We therefore compare the supervised additional attention layer to our baseline without this additional layer. Supervising the additional attention layer significantly improves in-distribution performance with further improvements on SNLI-hard and MNLI (see the top half of Table \ref{significance_testing}). While these results are also promising, we focus the remainder of the paper on supervising existing attention layers where we see greater improvements.

The in-distribution benefits from training with the explanations contrast with previous work on model robustness, with most work involving a trade-off between robustness and in-distribution performance \citep{sanh2020learning, karimi-mahabadi-etal-2020-end, belinkov-etal-2019-dont}. While some prior work retains in-distribution performance \citep{utama2020mind}, we find that training with explanations improves both in-distribution and out-of-distribution performance. 

\subsection{Experiments with DeBERTa}

We evaluate the effect of training with explanations for DeBERTa, assessing whether the human explanations can improve even more powerful NLI models. We find that DeBERTa itself achieves 92.59\% accuracy, outperforming previous state of the art results on SNLI \cite{zhang2020semanticsaware, pilault2021conditionally, sun2020selfexplaining}.
Combining the human explanations with DeBERTa provides a further statistically significant improvement for in-distribution performance, with the supervised model achieving 92.69\% performance, a new state of the art result for SNLI. While the absolute improvement is small (0.1\% for DeBERTa compared to 0.40\% for BERT), it is more challenging to achieve as the potential room for improvement has also decreased. 

\subsection{Comparing Results with Prior Work}

Our approach supervising existing model attention layers outperforms previously reported improvements, increasing SNLI performance by 0.40\%.  This compares to LIREx \citep{zhao2020lirex} which reported a 0.32\% improvement in SNLI accuracy when training with a pipeline of three RoBERTa models and a GPT2 model. We recreate this result (LIREx-adapted), replacing the RoBERTa models in the pipeline with BERT models, then compare it to our BERT baseline (Table \ref{benchmarking_results}). As previous work using e-InferSent \citep{camburu2018esnli}, TextCat \cite{hase2021models} and NILE \citep{kumar-talukdar-2020-nile} found no significant improvements using the explanations, we do not recreate these baselines.
\begin{table}[ht!]
\begin{center}
\begin{tabular}{rcccc}
\toprule
{\bf Explanation type} & {\bf Dev accuracy} & $\Delta$ \\ 
\midrule
Baseline & 89.89 & \\
Free text explanation & 90.35 & +0.46 \\
Highlighted words & 90.41 &  +0.52 \\
\midrule
Combined performance & \textbf{90.46} &  +0.57 \\
\bottomrule
\end{tabular}
\end{center}
\caption{Performance improvements were observed either when using free-text explanations or highlighted words, with the greatest improvements using a combination of these. Dev. accuracy is an average from 5 random seeds.}
\label{types_of_explanation}
\end{table}
We find that LIREx-adapted has the largest improvement compared to the BERT baseline (+1.02\%). This is unsurprising given that LIREx consists of a pipeline of four separate models, with a total of 453m parameters, compared to 109m parameters in the BERT baseline. In contrast, our approach of supervising an existing attention layer does not increase the number of parameters. LIREx-adapted also has a substantially lower performance than our DeBERTa model supervised with the explanations (90.79\% for SNLI-test compared to 92.69\%), despite using more parameters (453m compared to 409m).

No previous work has shown out-of-distribution improvements from training with the explanations, and this continues to be the case with LIREx-adapted: the SNLI improvements for LIREx-adapted are accompanied by a fall in MNLI performance (-0.85), and almost no change in the SNLI-hard performance (Table \ref{benchmarking_results}).

We additionally show that adapting the approach presented by \citet{pruthi2020evaluating} for NLI can also improve performance, with improvements across SNLI, MNLI and SNLI-hard. However, while improvements on MNLI are similar to our approach, improvements in SNLI-test are about half of the improvements we observed.

\subsection{Choosing Which Explanations to Use and Which Heads to Supervise}

We investigate different ways to use the e-SNLI explanations, assessing whether it is better to use the free-text explanations or the highlighted words. We also assess which attention heads should be supervised during training.  

\begin{table*}[ht!]
\begin{center}
\begin{tabular}{rcccccc}
\toprule
%
%
%

 & \multicolumn{3}{c}{Premise} & \multicolumn{3}{c}{Hypothesis} \\
 & P & R & F1 & P & R & F1\\ 

\midrule

Supervised LSTM-CRF \cite{thorne-etal-2019-generating} & 86.91 & 40.98 & 55.70 & 81.16 & 54.79 & 65.41 \\ 
Unsupervised attention threshold \citep{thorne-etal-2019-generating} & 19.23 & 26.21 & 22.18 & 53.38 & 62.97 & 57.78 \\
LIME \citep{thorne-etal-2019-generating} & 60.56 & 48.28 & 53.72 & 57.04 & 66.92 & 61.58 \\
SE-NLI \citep{10.1145/3418052} & 52.5 & 72.6 & \textbf{60.9} & 49.2 & 100.0 & 66.0 \\
\midrule
 Baseline, with no supervision & 0.51 & 0.01 & 0.03 & 43.32 & 58.65 & 49.83 \\ 
 Ours (existing attention) & 55.20 & 58.60 & 56.85 & 61.48 & 78.96 & \textbf{69.13} \\
\bottomrule
\end{tabular}

\end{center}
\caption{Precision, recall and F1 scores from token level predictions, using average attention values from 3 supervised attention heads. This is compared to a supervised LSTM-CRF model, LIME, SE-NLI, and the unsupervised attention approach.} 
\label{token_predictions}
\end{table*} 
\begin{figure}[ht!]
    \includegraphics[width=\columnwidth]{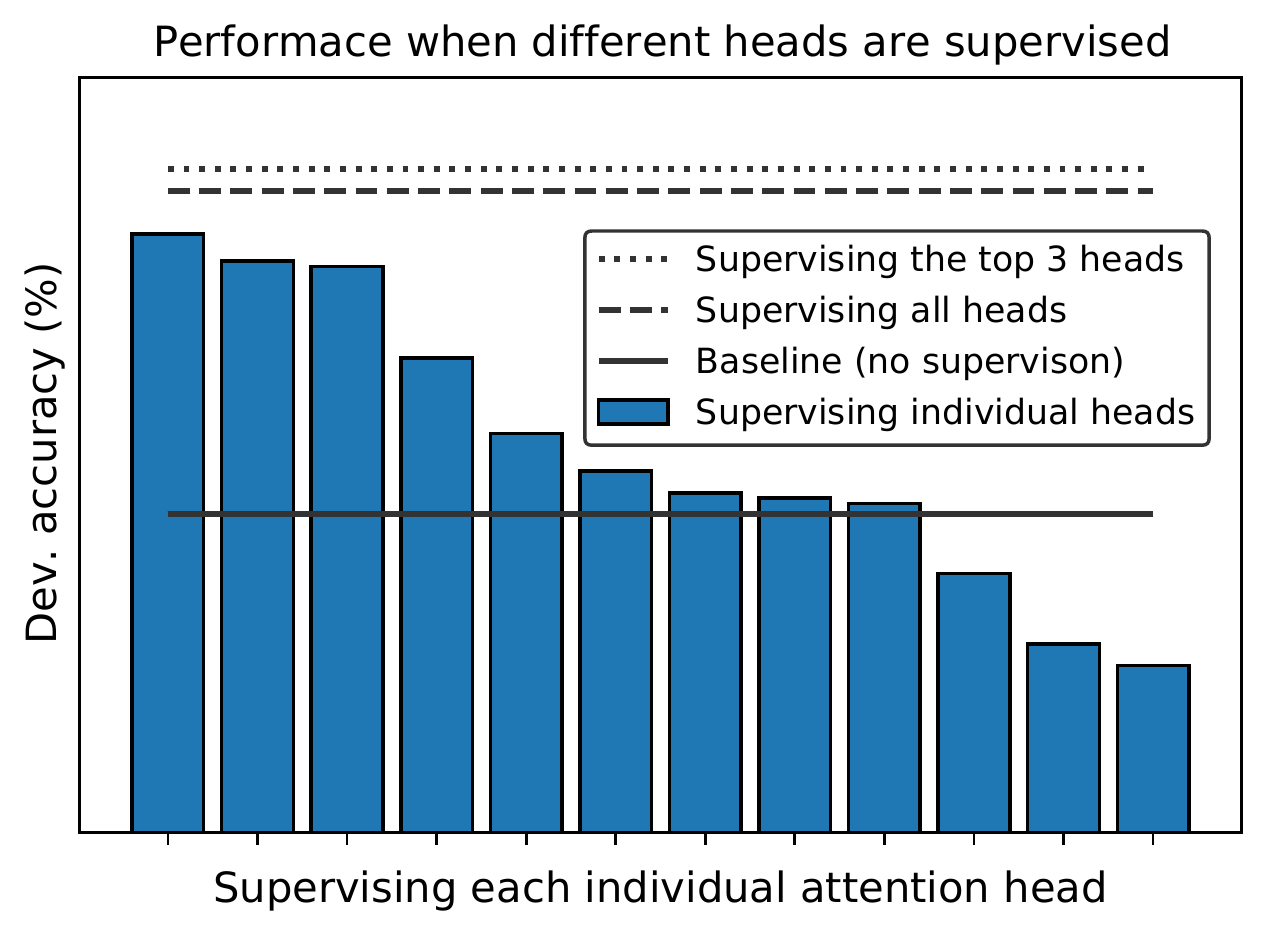}
    \caption{Accuracy when supervising each of the attention heads in turn, compared to the baseline with no supervision, supervising all heads and supervising the top 3 heads.} \label{individual_heads}
\label{which_head}
\end{figure}

We find the best performance when combining both the free text explanations and the highlighted words within e-SNLI, taking an average of their attention distributions, $D_{freetext}$ and $D_{highlights}$  (see Table \ref{types_of_explanation}). When there are only words highlighted in the hypothesis for $D_{highlights}$, the attention is supervised using $D_{freetext}$, encouraging the model to pay attention to both sentences. 

While we show that supervising all attention heads results in performance improvements (Figure \ref{which_head}), we find the best performance when only supervising 3 attention heads. This demonstrates how the additional supervision is only helpful for some attention heads, depending on the role of that specific head. 
Multi-head attention is designed to allow each head to perform a different function, therefore supervising all of them in the same direction can potentially have adverse effects.
Figure \ref{which_head} shows that the top 3 heads clearly performed better than the remaining heads when supervised individually, suggesting why this was the optimal number.

\section{Analysis}

\subsection{Token Level Classification}
To measure how successful the supervised heads are at identifying words in the human explanations, we consider the task of predicting which words appear in the highlighted explanations. The token-level classification is achieved by applying a threshold to the supervised attention weights, predicting whether a token is highlighted or not within e-SNLI. Unlike \citet{thorne-etal-2019-generating},  \citet{DBLP:conf/naacl/ReiS18} and \citet{bujel2021zero}, we apply the token level thresholds to the normalised attention weights instead of the unnormalised weights, finding that this improves performance.

The model's token level predictions outperform a LSTM-CRF model jointly supervised for NLI and the token level task \cite{thorne-etal-2019-generating, lample-etal-2016-neural} (see Table \ref{token_predictions}). We also compare this to an unsupervised approach using attention weights to make predictions \cite{thorne-etal-2019-generating}, LIME \cite{thorne-etal-2019-generating, ribeiro-etal-2016-trust} and a perturbation-based self-explanation approach \cite{10.1145/3418052}. The hypothesis F1 score for our approach is higher than previous baselines, with an improvement of 3.1 points. While \citet{10.1145/3418052} find a higher F1 score for the premise, their work focused on improving the token level performance and did not improve the overall NLI task.

\subsection{Understanding the Changes in Attention}
\begin{figure*}[ht!]
    \includegraphics[width=\textwidth, height=76pt]{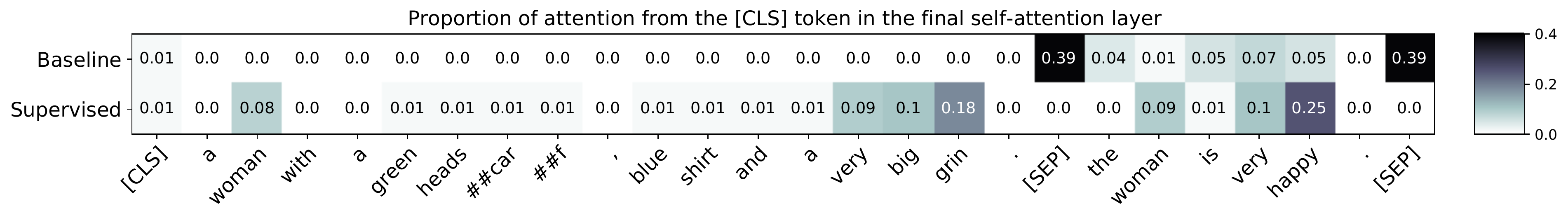}
    \includegraphics[width=\textwidth, height=76pt]{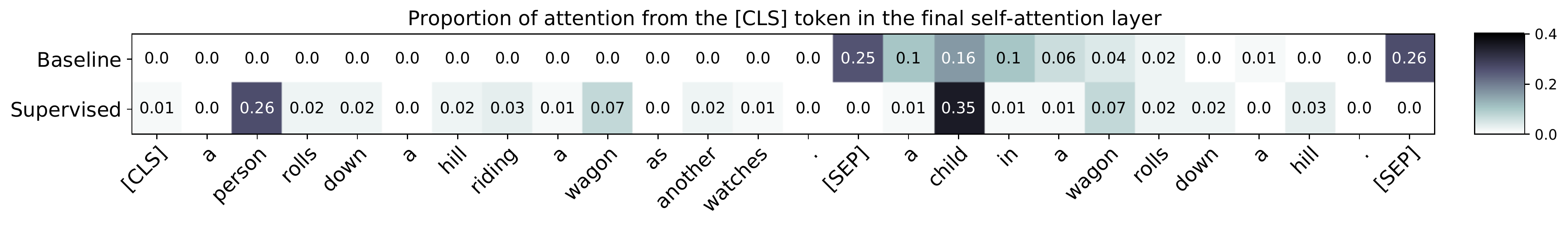}

    \caption{Average attention from the [CLS] token in the baseline and when we are supervising each attention head. Both models incorrectly predicted the first example as being neutral. The second example was correctly labeled by the supervised model (neutral), while the baseline model incorrectly predicted contradiction. The e-SNLI free-text explanations for the sentences include: `One must be happy in order to have a big grin' and `Just because it is a person does not mean it is a child'.}
    \label{example_sentences}
\end{figure*}
To understand how the attention behaviour changes in our supervised model, we analyse the final [CLS] token attention compared to the baseline. The premise and the 1st [SEP] token only account for 22.86\% of attention in the baseline, compared to 50.89\% when supervising 12 heads. This highlights how the supervised model more evenly considers both the premise and hypothesis compared to the baseline.

Even in the earlier attention layers which were not directly supervised, more attention is paid to the premise in the supervised model (with 31.1\% of attention in the baseline for the previous layer, compared to 54.2\% with supervision). The increased focus on the premise may explain why performance is substantially better for SNLI-hard, a challenge set created from examples that a hypothesis-only model misclassified. Surprisingly, if we supervise only 3 heads in the top layer, lower layers attend to the premise to the same extent (with 54.8\% of attention in the previous layer when supervising only 3 heads). This supports our decision to supervise fewer heads.
\begin{table}[ht!]
\begin{center}
\begin{tabular}{rcccc}
\toprule
%
%
%
{\bf PoS Tag} & 12 heads  & 3 heads & Baseline \\ 
\midrule
Noun &\textbf{ 54.3} & 43.5 & 28.1 \\
Verb &\textbf{ 20.4 } & 18.2 & 14.3 \\
Adjective & \textbf{8.9} & 8.3 & 5.2 \\
Adposition & 4.1 & 5.0 &\textbf{7.8} \\
Determiner & 3.4 & 6.0 &\textbf{14.3} \\
Punctuation & 0.9 & 7.7 & \textbf{14.2} \\
Auxiliary & 0.9 & 3.1 &\textbf{8.2} \\
Other & 7.1 & 8.2 & 7.9
\\
\bottomrule
\end{tabular}

\end{center}
\caption{Percentage of attention across 5 seeds from the [CLS] token to tokens corresponding to different PoS tags.} 
\label{pos_tagging}
\end{table}

\begin{table}[t!]
\begin{center}
\begin{tabular}{ccccc}
\toprule
%
%
%
\multicolumn{2}{c}{\textbf{Baseline}} & \multicolumn{2}{c}{\textbf{Supervised}} \\
Words & \% & Words & \%  \\ 
\midrule
. & 18.0 & man & 2.7  \\
a & 5.2 & outside & 2.5\\
is & 4.0 & woman & 1.7 \\
are & 2.6 & people & 1.7 \\
the & 2.5 & sitting & 1.5\\
\bottomrule
\end{tabular}
\end{center}
\caption{Frequency in which each word is the most attended to token in a sentence pair across 5 random seeds.}
\label{most_attended_words}
\end{table} 
\subsection{Words Receiving Most Attention}

In the supervised model, the words that receive the most attention are often nouns such as \textit{man}, \textit{woman}, or \textit{people} (Table \ref{most_attended_words}) which are the subjects of many sentences. Nouns are frequently used in the explanations, making up 46\% of the highlighted words. On the other hand, stop-words are often attended to in the baseline, along with full-stops which may be a form of null attention \cite{vig-belinkov-2019-analyzing}.
More generally, using a SpaCy\footnote{https://spacy.io} Part of Speech tagger, after supervision we see less attention paid to punctuation, determiners and adposition words, while more attention is paid to nouns, verbs and adjectives (Table \ref{pos_tagging}). 

An analysis of the attention behaviour shows that the supervised model consistently attends to the most important words for the task, which is often not the case for the baseline model. In Figure \ref{example_sentences}, for each example the supervised model identifies the most important words in both the premise and the hypothesis. In the first sentence pair it attends to the word `grin' in the premise and `happy' in the hypothesis. In the second example, the supervised model identifies that the `person' in the premise and `child' in the hypothesis are the most important words. 

Unlike the baseline, which mostly attends to the hypothesis and special tokens, the supervised model attends to words in the premise. As a result, the behaviour of the supervised model is more interpretable for NLI, where the class depends on the interaction between the two sentences. 

\section{Conclusion}
Motivated by improving the robustness of NLI models based on human behaviour, we introduce a simple but effective approach that helps models learn from human explanations. 
We find the best performance when supervising a model's existing self-attention weights, encouraging more attention to be paid to words that are important in human explanations.
Unlike prior work incorporating human explanations, our approach improves out-of-distribution performance alongside in-distribution performance, achieving a new state of the art result when combined with a DeBERTa model.
Our supervised models have more interpretable attention weights and focus more on the most important words in each sentence, mostly nouns, verbs and adjectives. This contrasts with the baseline model that attends more to special tokens, stop-words and punctuation. The result is a model that attends to words humans believe are important, creating more robust and better performing NLI models.

\section{Acknowledgments}

This research was partly supported by the ISRAEL SCIENCE FOUNDATION (grant No. 448/20). Y.B. was supported by an Azrieli Foundation Early Career Faculty Fellowship and by the Viterbi Fellowship in the Center for Computer Engineering at the Technion. We would like to thank the authors of the e-SNLI dataset for creating this excellent resource, and we also thank the LAMA reading group at Imperial for their feedback and encouragement.


\bibliography{anthology}


\end{document}